\documentclass[letterpaper, 10 pt, conference]{ieeeconf}  
\IEEEoverridecommandlockouts                              
\usepackage{graphics} 
\usepackage{epsfig} 
\usepackage{mathptmx} 
\usepackage{times} 
\usepackage{amsmath} 
\usepackage{amssymb}  
\usepackage{multirow}
\usepackage{mathrsfs}
\usepackage{subcaption}
\title{\LARGE \bf
Enhancing Depression Detection with Chain-of-Thought Prompting: From Emotion to Reasoning Using Large Language Models
}

\author{Shiyu Teng$^{1}$, Jiaqing Liu$^{1}$, Rahul~Kumar~Jain$^{1}$, Shurong Chai$^{1}$, Ruibo Hou$^{1}$,  Tomoko Tateyama$^{2}$,  Lanfen Lin$^{3}$ \\and *Yen-wei Chen$^{1}$
\thanks{$^{1}$Shiyu Teng, Shurong Chai, Ruibo Hou, Rahul~Kumar~Jain, Jiaqing Liu and Yen-wei Chen are with the College of Information Science and Engineering, Ritsumeikan  University, Osaka, Japan.}
\thanks{$^{2}$Tomoko Tateyama is with Department of Intelligent Information Engineering, Fujita Health University, Japan}%
\thanks{$^{3}$Lanfen Lin is with the College of Computer Science and Technology, Zhejiang University, Hangzhou, China}
\thanks{*Corresponding Authors: Yen-Wei Chen (chen@is.ritsumei.ac.jp)
}%
}

\begin{document}

\maketitle
\thispagestyle{empty}
\pagestyle{empty}

\begin{abstract}
Depression is one of the leading causes of disability worldwide, posing a severe burden on individuals, healthcare systems, and society at large. Recent advancements in Large Language Models (LLMs) have shown promise in addressing mental health challenges, including the detection of depression through text-based analysis. However, current LLM-based methods often struggle with nuanced symptom identification and lack a transparent, step-by-step reasoning process, making it difficult to accurately classify and explain mental health conditions. To address these challenges, we propose a Chain-of-Thought Prompting approach that enhances both the performance and interpretability of LLM-based depression detection. Our method breaks down the detection process into four stages: (1) sentiment analysis, (2) binary depression classification, (3) identification of underlying causes, and (4) assessment of severity. By guiding the model through these structured reasoning steps, we improve interpretability and reduce the risk of overlooking subtle clinical indicators. We validate our method on the E-DAIC dataset, where we test multiple state-of-the-art large language models. Experimental results indicate that our Chain-of-Thought Prompting technique yields superior performance in both classification accuracy and the granularity of diagnostic insights, compared to baseline approaches. 
\end{abstract}

\section{INTRODUCTION\label{introduction}}

Depression is one of the most prevalent and severe mental health disorders worldwide, posing significant challenges to both individuals and society \cite{santomauro2021global}. According to the World Health Organization, it is a leading cause of disability and a major contributor to the global burden of disease. The high prevalence of depression and its strong link to suicide highlight the urgent need for early detection and intervention \cite{who_depression}. In recent years, advancements in artificial intelligence have opened new possibilities for identifying depression through linguistic patterns. Large Language Models (LLMs), such as ChatGPT \cite{achiam2023gpt} and DeepSeek \cite{deepseek}, have shown strong capabilities in capturing semantic and contextual nuances. By leveraging these models’ deep understanding of language, early signs of depression can be detected more accurately, facilitating timely psychological support and treatment.

Despite their potential, current LLM-based approaches to mental health analysis have notable limitations. Most existing methods rely on direct classification frameworks, mapping text to diagnostic labels or inferred reasoning outcomes \cite{Yang_2024}. However, this end-to-end approach presents three key challenges in clinical practice. First, these models lack explicit reasoning about symptoms, making their decision process less interpretable and failing to meet the medical standard of auditability \cite{amann2020explainability}. Second, holistic text processing often overlooks crucial depressive symptoms that require structured linguistic analysis. For example, anhedonia (loss of pleasure) is frequently expressed through negated positive affect rather than direct negative statements. Clinical linguistics studies indicate that phrases like “I try to enjoy hobbies but feel nothing” or “Family gatherings should be fun, yet they feel meaningless” contain two diagnostic cues: (1) an implicit expectations of positive experiences (“enjoy,” “fun”), and (2) contextual negation (“but feel nothing,” “yet meaningless”). Identifying such patterns requires structured evaluation, yet current LLMs struggle with this task \cite{al2018absolute}. Third, existing models conflate symptom identification with severity assessment, whereas in clinical practice, these should be separate processes. Symptom detection is best handled as a binary classification task, while severity assessment requires both multi-class classification and regression \cite{kroenke2001phq}. These limitations reduce diagnostic precision and hinder clinical adoption.

\begin{figure}[t]
    \centering
    \begin{subfigure}{0.23\textwidth}
        \centering
        \includegraphics[width=\linewidth]{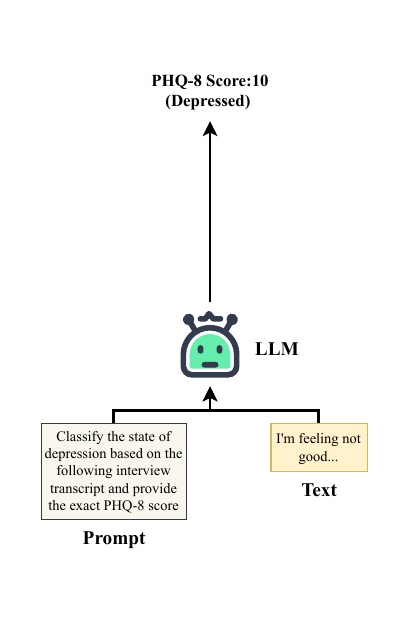}
        \caption{Standard Prompting}
        \label{fig:image1}
    \end{subfigure}
    \hfill
    \begin{subfigure}{0.23\textwidth}
        \centering
        \includegraphics[width=\linewidth]{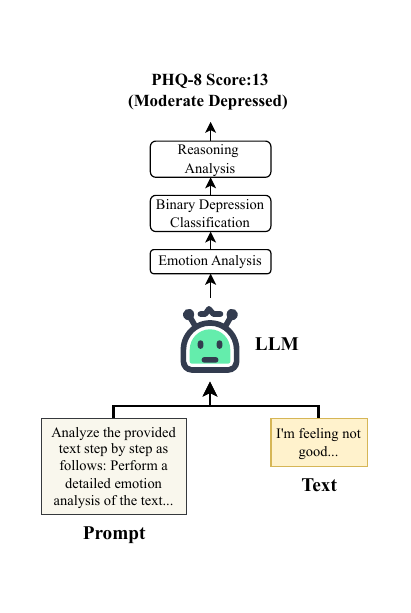}
        \caption{Chain-of-Thought (Ours)}
        \label{fig:image2}
    \end{subfigure}
    \caption{Comparison of Standard Prompting and Chain-of-Thought Prompting for Depression Assessment.}
    \label{fig:two_images}
\end{figure}

Recent advancements in Chain-of-Thought (CoT) reasoning have significantly improved LLMs’ ability to analyze complex mental health assessments \cite{wei2023chainofthoughtpromptingelicitsreasoning}. Models such as GPT-o1 \cite{achiam2023gpt}, DeepSeek R1 \cite{deepseek}, and QwQ \cite{qwen2.5} have demonstrated notable improvements in capturing nuanced psychological cues by breaking down reasoning into explicit, interpretable steps. This structured approach aligns closely with clinical diagnostic workflows, where mental health professionals assess depression through systematic evaluations rather than direct label assignment. To address the aforementioned challenges, we propose a CoT prompting strategy that transitions from emotion analysis to causal reasoning, mimicking both CoT-based LLM reasoning and clinical diagnostic procedures. Our method follows a structured process: first, the model identifies emotional expressions indicative of depression, such as anhedonia or hopelessness. Next, it performs a binary classification to determine whether depressive tendencies are present. Based on the detected emotional cues, the model then conducts causal reasoning to explore potential psychological or social factors contributing to the condition. Finally, it combines all extracted information to assess depression severity, ensuring a structured and interpretable evaluation aligned with clinical standards. We evaluated our emotion-to-reasoning CoT framework on the E-DAIC dataset \cite{ringeval2019avec}, which consists of clinical interviews annotated with depression severity levels. Our experiments compared three configurations: (1) standard LLMs (e.g., GPT-4o \cite{achiam2023gpt}, Qwen2.5-Max \cite{qwen2.5}), (2) CoT-enhanced LLMs (e.g., DeepSeek-R1 \cite{deepseek}, GPT-o1 \cite{achiam2023gpt}, GPT-o3-mini \cite{achiam2023gpt}, QWQ \cite{qwen2.5}), and (3) our structured CoT approach applied to both standard and CoT-enhanced LLMs. Results indicate that our structured CoT approach not only achieves higher accuracy but also provides richer diagnostic explanations compared to both standard and CoT-enhanced LLMs.

Our contributions can be summarized as follows:
\begin{itemize}
\item We conduct comprehensive benchmarking of several state-of-the-art LLMs for depression detection, revealing significant performance variations and establishing crucial baselines for model selection in clinical applications.

     \item We propose a novel CoT prompting strategy that decomposes the depression detection task into four structured stages: sentiment analysis, binary classification, underlying cause identification, and severity assessment. This structured approach not only mimics the clinical diagnostic process but also enhances interpretability and transparency by explicitly guiding the model through each reasoning step.
    \item Experimental results on the E-DAIC dataset demonstrate that our CoT prompting technique outperforms traditional methods in both classification accuracy and the granularity of diagnostic insights, highlighting its potential for clinical applications in mental health assessments.
\end{itemize}

\begin{figure*}[!t]
\begin{center}
\includegraphics[width=0.8\linewidth]{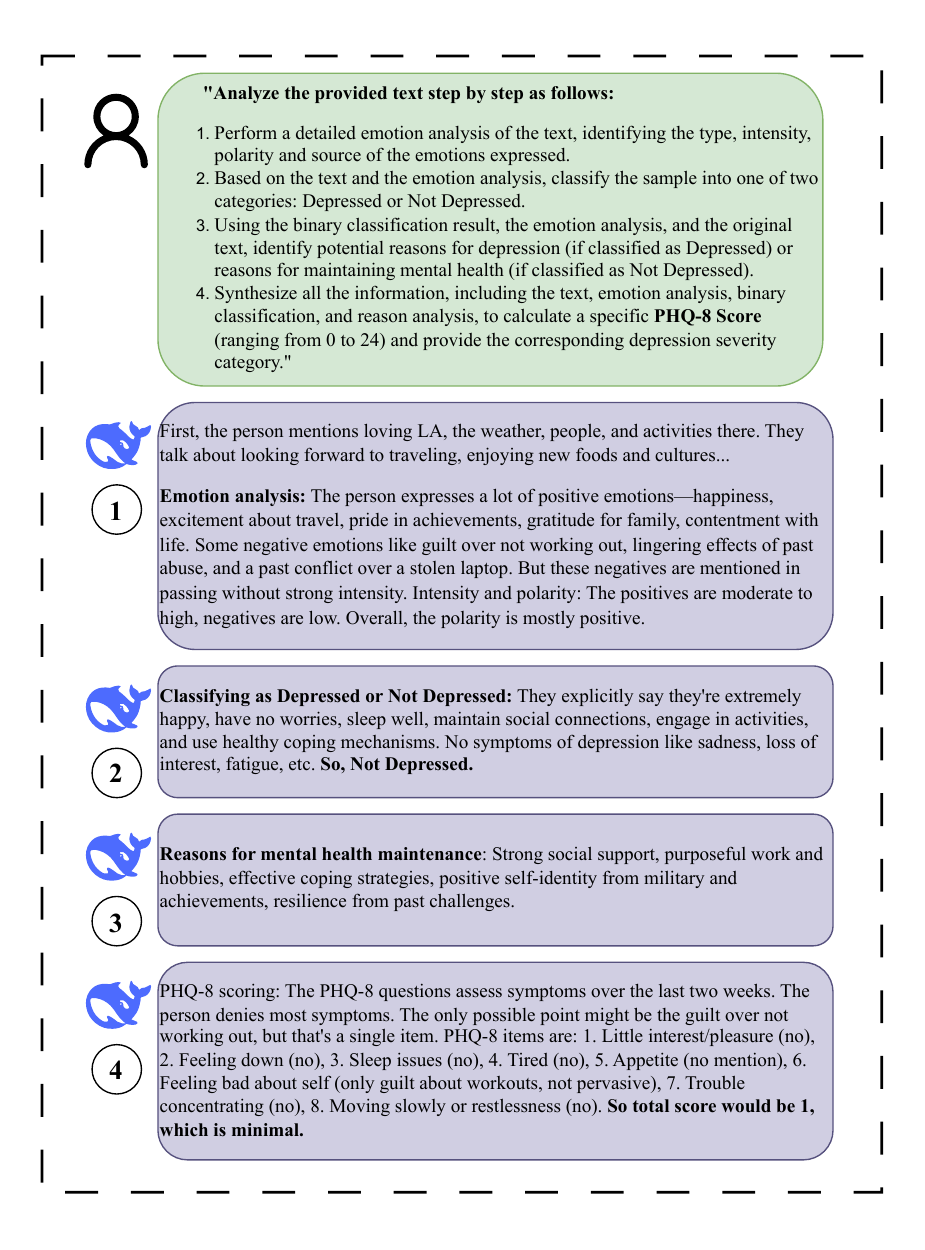}
\end{center}
\caption{Our CoT prompting framework with four-stage clinical reasoning process. 
}
\label{fig:overview}
\end{figure*}

\section{METHODOLOGY\label{methodology}}
Our Chain-of-Thought (CoT) prompting framework decomposes depression detection into four sequential reasoning stages, as shown in Figure~\ref{fig:overview}. The pipeline processes input text through:

\begin{enumerate}
\item Emotion Analysis: Identifying emotion type, intensity, polarity, and source.
\item Binary Classification: Determining whether the individual is classified as Depressed or Not Depressed.
\item Reasoning Analysis: Identifying factors contributing to depression or mental well-being.
\item Severity Assessment: Calculating the PHQ-8 Score and assessing depression severity.
\end{enumerate}

\subsection{Stage 1: Emotion Analysis}
The initial stage employs a structured approach to extract detailed emotional signals. This includes identifying the type of emotion (e.g., sadness, guilt, hope), its intensity (low, medium, or high), its polarity (positive, negative, or neutral), and its source (internal thoughts, external events, relationships, or health). By capturing these emotional indicators, including subtle expressions such as anhedonia through negated positives (e.g., Family gatherings should be fun, yet...), the system can provide a comprehensive emotional profile of the input text.

\subsection{Stage 2: Binary Classification}
Using the emotional features extracted in Stage 1, we prompt the LLM to determine depression likelihood:

\begin{equation}
P(depression|s) = f_{\theta}(s, E_{\text{emotion}}),
\end{equation}
where $s$ represents the input text and $E_{\text{emotion}}$ represents extracted emotional features. The classification follows PHQ-8 clinical guidelines \cite{kroenke2009phq} to categorize the individual as Depressed or Not Depressed.

\subsection{Stage 3: Reasoning Analysis}
For individuals classified as Depressed, the system activates causal reasoning is activated to determine underlying contributing factors. Conversely, for non-depressed cases, the system identifies protective factors contributing to mental well-being.

If the individual is classified as Depressed, potential contributing factors are identified across multiple dimensions:
\begin{itemize}
\item Social factors such as isolation, conflict, or lack of support
\item Biological factors including sleep disturbances, appetite changes, and fatigue
\item Psychological factors like guilt, worthlessness, and negative self-perception
\item Functional impairment affecting work, relationships, and daily activities
\end{itemize}
A ranked list of depressive factors is generated, supported by textual evidence.

If the individual is classified as Not Depressed, the system identifies positive influences and protective factors, such as:
\begin{itemize}
\item Social support systems and fulfilling relationships
\item Psychological resilience, coping mechanisms, and self-esteem
\item Healthy habits including consistent sleep, balanced nutrition, and exercise
\end{itemize}
A ranked list of these positive influences is provided with supporting evidence.

\subsection{Stage 4: Severity Assessment}
Using insights from the previous stages, we determine depression severity based on PHQ-8 scoring. The PHQ-8 score ranges from 0 to 24 and is categorized as follows:
\begin{itemize}
\item Minimal: 0-4
\item Mild: 5-9
\item Moderate: 10-14
\item Moderately Severe: 15-19
\item Severe: 20-24
\end{itemize}
The final output provides both the PHQ-8 score and the corresponding depression severity category, ensuring a quantitative evaluation aligned with clinical standards for diagnosis and treatment assessment.

\section{EXPERIMENTS\label{EXPERIMENTS}}

\subsection{Dataset \label{subsec:dataset}}

The Extended Distress Analysis Interview Corpus (E-DAIC) dataset \cite{ringeval2019avec} consists of audiovisual recordings of clinical interviews designed for depression detection, where a virtual agent conducts interviews to minimize human interaction bias. It includes text, acoustic, and visual modalities, with samples annotated using PHQ-8 scores (0-24) to indicate depression severity. The dataset is divided into 163 training, 56 validation, and 56 test samples, as detailed in Table \ref{table:distribution}. In our experiments, we utilize only the text modality, focusing on linguistic cues for depression assessment.

\begin{table}[ht]
\centering
\caption{Distribution of training, validation, and test sets.}
\begin{tabular}{llccc}
\hline
\textbf{Task} & \textbf{Category} & \textbf{Train} & \textbf{Val} & \textbf{Test} \\ \hline
Regression & - & 163 & 56 & 56 \\ \hline
\multirow{5}{*}{Classification} 
& Minimal (0-4) & 77 & 26 & 26 \\ 
& Mild (5-9) & 36 & 15 & 15 \\ 
& Moderate (10-14) & 26 & 8 & 8 \\ 
& Moderately Severe (15-19) & 17 & 6 & 6 \\ 
& Severe (20-24) & 7 & 1 & 1 \\ \hline
\end{tabular}
\label{table:distribution}
\end{table}

\subsection{Evaluation Metrics}

In this study, we adopt two primary metrics to assess the performance of our regression model in predicting depression severity: the concordance correlation coefficient (CCC) and the mean absolute error (MAE). The CCC \cite{lawrence1989concordance} is a prevalent measure in depression detection research, as it evaluates how well the predicted scores align with the true depression severity ratings. It is defined as:

\begin{equation} CCC = \frac{2 s_{\hat{y} y}}{s_{\hat{y}}^2 + s_y^2 + (\bar{\hat{y}} - \bar{y})^2}, \label{eq:CCC} \end{equation}

where $\bar{\hat{y}}$ and $\bar{y}$ are the mean values of the predictions and the ground truth, respectively; $s_{\hat{y}}^2$ and $s_y^2$ denote their variances; and $s_{\hat{y} y}$ represents the covariance between them. The CCC value ranges between -1 and 1, with -1 indicating complete disagreement and 1 indicating perfect agreement.

Furthermore, we use the mean absolute error (MAE) to quantify the average prediction error. The MAE is given by:

\begin{equation} MAE = \frac{1}{N} \sum_{i=1}^{N} \left| \hat{y}_i - y_i \right|, \label{eq:MAE} \end{equation}

where $N$ is the total number of samples, $\hat{y}_i$ is the predicted score for the $i$th sample, and $y_i$ is the corresponding true score. 

\subsection{Experiment Results}

Table \ref{table:comparison} presents the performance comparison of our proposed CoT prompting approach with existing state-of-the-art methods on the E-DAIC test set. The results are evaluated using the CCC and MAE, where higher CCC and lower MAE values indicate better performance. Among the traditional multimodal approaches, CubeMLP \cite{Cubemlp} and MIMRL \cite{sunMutual} achieved competitive results, with CCC scores of 0.583 and 0.580, respectively, and MAE values of 4.37 and 4.36, respectively. These methods leverage advanced deep learning architectures for multimodal fusion but still exhibit limitations in capturing nuanced linguistic cues associated with depression.

In contrast, LLMs demonstrate a significant improvement in performance compared to traditional multimodal methods, particularly when integrated with our CoT prompting strategy. Among the standard LLMs, DeepSeek V3 \cite{deepseek} achieves a CCC of 0.622 and an MAE of 4.15, already surpassing previous multimodal deep learning approaches. Similarly, Qwen2.5-Max \cite{qwen2.5} further improves upon this, achieving a CCC of 0.637 and an MAE of 4.07. GPT 4o \cite{achiam2023gpt} exhibits the highest performance among standard LLMs, reaching a CCC of 0.732 and the lowest MAE of 3.37.

When applied to LLMs with inherent CoT reasoning capabilities, our prompting strategy further enhances their interpretability and diagnostic comprehensiveness. For instance, GPT o3-mini \cite{achiam2023gpt} and GPT o1-preview \cite{achiam2023gpt} achieve CCC values of 0.677 and 0.690, respectively, demonstrating improved step-by-step reasoning and classification accuracy. Additionally, QwQ-32b-preview \cite{qwen2.5} and DeepSeek R1 \cite{deepseek} further improve CoT-based depression assessment, with DeepSeek R1 achieving a CCC of 0.708 and an MAE of 3.49, indicating the highest precision in depression severity estimation among the CoT-enhanced models. These results highlight the versatility of our Chain-of-Thought prompting approach, which benefits LLMs. Our method enables standard LLMs to adopt structured reasoning capabilities similar to CoT-enhanced models, leading to improved performance. For models that already possess CoT functionality, our prompting strategy improves their reasoning process. It helps them analyze emotional and causal factors in more detail. This results in a more comprehensive assessment of depression. By systematically guiding models through emotion analysis, depression classification, causal reasoning, and severity assessment, our method not only improves accuracy but also enhances interpretability, making it well-suited for clinical mental health assessment applications.

\begin{table}[ht]
\centering
\caption{The results on the test set of the E-DAIC dataset.}
\begin{tabular}{cccc}
\hline
Category & Method & CCC $\uparrow$ & MAE $\downarrow$ \\
\hline
\multirow{5}{*}{Multimodal Methods} 
& AFT \cite{sun2021multi} & 0.443 & 5.66 \\
& Teng \emph{et al.}\cite{teng2024icce} & 0.466 & 5.21 \\
& Tensorformer \cite{sun2022tensorformer} & 0.493 & 5.13 \\
& STMCAT \cite{tengembc} & 0.507 & 4.77 \\
& CubeMLP \cite{Cubemlp} & 0.583 & 4.37 \\ 
& MIMRL \cite{sunMutual} & 0.580 & 4.36 \\ 
\hline
\multirow{7}{*}{\shortstack{LLMs with our\\ COT prompts}}
& DeepSeek V3 \cite{deepseek}  & 0.622 & 4.15 \\ 
& Qwen2.5-Max \cite{qwen2.5}  & 0.637 & 4.07 \\ 
& GPT o3-mini \cite{achiam2023gpt} & 0.677 & 3.68 \\ 
& GPT o1-preview \cite{achiam2023gpt} & 0.690 & 3.60 \\ 
& QwQ-32b-preview \cite{qwen2.5} & 0.705 & 3.55 \\ 
& DeepSeek R1 \cite{deepseek} & 0.708 & 3.49 \\ 
& GPT 4o \cite{achiam2023gpt} & \textbf{0.732} & \textbf{3.37} \\ 
\hline
\end{tabular}
\label{table:comparison}
\end{table}


\subsection{Ablation Study}

Table \ref{table:ablation} presents the results of our ablation studies, which evaluate the impact of incorporating our CoT prompting strategy on various LLMs. The table is divided into two groups: the upper group consists of standard LLMs that do not possess inherent CoT capability, while the lower group comprises models that already have inherent CoT functionality.

For the standard LLMs, our results clearly indicate that integrating CoT prompting substantially improves performance. Specifically, for Qwen2.5-Max \cite{qwen2.5}, the application of CoT prompting increases the CCC from 0.55 to 0.637 and reduces the MAE from 4.33 to 4.07. Similarly, GPT 4o \cite{achiam2023gpt} exhibits an improvement in CCC from 0.696 to 0.732 and a reduction in MAE from 3.47 to 3.37 when CoT prompting is applied. These improvements suggest that our CoT prompting approach effectively equips standard LLMs with enhanced reasoning capabilities, leading to more precise depression severity estimation. 

For LLMs that already possess inherent CoT capabilities, our method further refines their performance. As shown in the lower group, GPT o3-mini \cite{achiam2023gpt} benefits from an increase in CCC from 0.625 to 0.677 and a decrease in MAE from 3.84 to 3.68 with CoT prompting. Similarly, QwQ-32b-preview \cite{qwen2.5} improves its CCC from 0.597 to 0.705 and reduces its MAE from 4.23 to 3.55. These findings demonstrate that even for models with built-in CoT reasoning, our structured prompting strategy can further enhance their reasoning depth and diagnostic comprehensiveness. 

Overall, these ablation studies confirm that our CoT prompting strategy significantly bolsters the performance of LLMs in depression assessment tasks. It not only endows standard LLMs with advanced reasoning capabilities but also refines the thought process of CoT-enhanced models, resulting in improved accuracy and interpretability in clinical mental health assessments.

\begin{table}[htbp]
\centering
\caption{Results of ablation studies. The upper group shows standard LLMs (without inherent CoT capability) while the lower group comprises LLMs with inherent CoT functionality.}
\begin{tabular}{l c | c c}
\hline
Model            & COT Prompting & CCC $\uparrow$ & MAE $\downarrow$ \\ \hline
Qwen2.5-Max \cite{qwen2.5}     &               & 0.55  & 4.33  \\
Qwen2.5-Max  \cite{qwen2.5}    & \checkmark    & 0.637 & 4.07  \\
GPT 4o    \cite{achiam2023gpt}       &               & 0.696 & 3.47  \\
GPT 4o   \cite{achiam2023gpt}         & \checkmark    & \textbf{0.732} & \textbf{3.37} \\ \hline
GPT o3-mini  \cite{achiam2023gpt}         &               & 0.625 & 3.84  \\
GPT o3-mini  \cite{achiam2023gpt}         & \checkmark    & 0.677 & 3.68  \\
QwQ-32b-preview \cite{qwen2.5}  &               & 0.597 & 4.23  \\
QwQ-32b-preview \cite{qwen2.5}  & \checkmark    & 0.705 & 3.55  \\ \hline
\end{tabular}
\label{table:ablation}
\end{table}

\section{CONCLUSION AND FUTURE WORK\label{conclusion}}

In this work, we presented a novel CoT prompting strategy to improve depression detection using LLMs. Our method enhances both performance and interpretability. We decompose the depression detection task into four structured stages: emotion analysis, binary classification, causal analysis, and severity assessment. This structured approach follows the clinical diagnostic process. It provides detailed, step-by-step reasoning. Experimental results on the E-DAIC dataset show that our CoT approach performs better than traditional multimodal methods and standard LLMs. Models integrated with our CoT prompting strategy achieve higher Concordance Correlation CCC and lower MAE. These results highlight the benefits of structured reasoning. It helps capture nuanced linguistic cues related to depression. Ablation studies confirm that our method improves standard LLMs by adding advanced reasoning capabilities. It also refines the reasoning process of CoT-enhanced models. This leads to better diagnostic accuracy and interpretability.

While our current study focuses solely on the text modality for depression assessment, recent advancements in large open-source models present exciting opportunities for incorporating multimodal data. In future work, we plan to extend our framework by integrating additional modalities such as audio and visual signals, which are known to carry rich emotional and behavioral cues. The fusion of multimodal information is expected to further enhance the accuracy and robustness of depression detection, providing a more comprehensive diagnostic tool that better reflects the complex nature of mental health.

\section*{ACKNOWLEDGMENT}
This work is supported in part by JSPS KAKENHI Grant Numbers JP23K16909, JST BOOST, Grant Number JPMJBS2428.

\bibliographystyle{IEEEtran}
\bibliography{refe.bib}

\end{document}